\documentclass[twocolumn,9pt]{IEEEtran}
\usepackage{preamble} 
\usepackage{textcomp}
\usepackage{subfigure}
\usepackage{courier}
\usepackage{stmaryrd}
\usetikzlibrary{chains,backgrounds}
\usetikzlibrary{intersections}

\usepackage{xstring} 
\usepackage{bbm}

\renewcommand{\captionN}[1]{\caption{\color{darkgray} \sffamily \fontsize{8}{10}\selectfont #1  }}

\tikzexternaldisable 
\newcommand{\Pp}[1][n]{\mathscr{P}^+_{#1}}
\newcommand{\Sp}[1][{\Sigma}]{\mathds{P}^+_{#1}}
\newcommand{\B}{\mathcal{B}_0}
\newcommand{\U}{\mathcal{U}_0}
\newcommand{\F}[1][\Sigma]{\mathscr{F}_{#1}}
\newcommand{\eqcl}[1][\phantom{x}]{\left [#1 \right ]}
\newcommand{\simn}[1][\mathcal{N}]{\sim_{{#1}}}

\newcommand{\nlz}[1]{\left \llbracket #1 \right \rrbracket}
\newcommand{\clx}[1]{\mathscr{C}\left ({#1}\right ) }
\newcommand{\cly}[1]{\mathscr{C}_\star\left ({#1} \right ) }
\newcommand{\pax}[1][]{(\Sigma,Q^{\mspace{.6mu}\mathclap{#1}},\delta^{\mspace{.6mu}\mathclap{#1}},\pitilde^{\mspace{.2mu}#1})}
\newcommand{\wax}[1][]{(\Sigma^\omega,\F,\mu)}
\newcommand{\paz}[1][]{(\Sigma,Q^{{#1}},\delta^{{#1}},\pitilde^{#1})}
\newcommand{\paw}[1][\star]{(\Sigma,Q^{{#1}},\delta,\pitilde)}
\newcommand{\pan}{\paz[\circ]}
\newcommand{\zr}[1][n]{\mathfrak{U}_{#1}}
\newcommand{\Zr}{\mathcal{Z}}

\newcommand{\CLX}[1]{\overline{{#1}}}
\DeclareMathOperator*{\argsup}{arg\,sup}

\vspace{-20pt}
\begin{document}  
\title{A Hilbert Space of Stationary Ergodic Processes}

\author{\IEEEauthorblockN{Ishanu Chattopadhyay\IEEEauthorrefmark{1}}
\IEEEauthorblockA{University of Chicago\\
Email: \IEEEauthorrefmark{1}ishanu@uchicago.edu}}
\maketitle


\begin{abstract}  
Identifying meaningful signal buried in noise is a problem of interest arising in 
diverse scenarios of data-driven modeling. We present here a theoretical framework for 
exploiting intrinsic geometry in data that resists noise corruption, and might be identifiable 
under severe obfuscation. Our approach is based on uncovering a valid complete inner product on the 
space of ergodic stationary finite valued processes, providing the latter with the structure of a Hilbert space on the real field.
This rigorous construction, based on non-standard generalizations of the notions of sum and scalar multiplication of 
finite dimensional probability vectors, allows us to meaningfully talk about ``angles'' between data streams and data sources, and,  
make precise the notion of orthogonal stochastic processes. In particular, the relative angles  appear to be 
preserved, and identifiable, under severe noise, and will be developed in future as the underlying principle 
for robust classification, clustering and unsupervised featurization algorithms.

\end{abstract}
\vspace{-15pt}
%
%
\allowdisplaybreaks{
\section{Preliminary Concepts}
\begin{defn}[Inner Product \& Inner Product Spaces]\label{defii}
An inner product on a real vector space $X$ is a function $\langle \cdots, \cdots \rangle :X \times X \to \mathbb{R}$, such that the following conditions are satisfied:
\begin{subequations}
\calign{
&\forall u,v,w \in X, \alpha \in \mathbb{R}, \langle u,\alpha(v+ w) \rangle = \alpha\langle u,v+w\rangle\notag \\ &\mspace{200mu} =\alpha\left ( \langle u,v\rangle +\langle u, w \rangle \right ) \tag{Bi-linearity} \\
&\forall v,w \in X, \langle v,w \rangle = \langle w,v \rangle \tag{Symmetry} \\
&\forall u \in X, \langle u,u \rangle \geqq 0, \textrm{ where } \langle u,u \rangle =0 \Rightarrow u=0  \tag{Positive Definiteness}
}\end{subequations}
\end{defn}
A vector space with an inner product is an inner product space. Note that an inner product necessarily  induces a norm, which in turn  induces a metric~\cite{kreyszig1989introductory,Rud88}.
\begin{defn}[Complete inner product space or Hilbert Space]
A complete inner product space, or a Hilbert space~\cite{Rud88}, is a Banach space with an inner product, $i.e.$, every Cauchy sequence in the space converges in the space. 
\end{defn}
\begin{notn}[Strictly Positive Probability Vectors]
For $n \in \mathbb{N}$, the space of  strictly positive probability vectors is defined as:
\cgather{
\Pp=\left \{ \wp \in \mathbb{R}^n : \forall i \ \wp_i >0, \sum_i \wp_i = 1 \right  \}
}
\end{notn}
\subsection{An Abelian Group on Probability Vectors}
$\Pp$ can be given the structure of an Abelian group~\cite{rotman1999introduction}, via the following binary operation: $\oplus: \Pp\times\Pp \to \Pp$~\cite{Chattopadhyay20140826}:
\mltlne{\label{eqsum}
\forall \wp,\wp' \in \Pp, \forall i \in \{1,\cdots,n\}, \\  \left ( \wp \oplus \wp' \right ) \bigg \vert_i \triangleq  \wp_i\wp_i'\left ( \sum_j\wp_j\wp'_j\right ) ^{-1}  
}
We denote $\oplus$ simply as $+$ in the sequel if there is no confusion. 
It is easy to see that we have the following properties (which makes $\Pp$ into an Abelian group, with $+$  as the group sum):
\begin{subequations}\cgather{
\forall \wp,\wp' \in \Pp, \wp+\wp' \in \Pp \\
\wp + \wp' = \wp' + \wp \\
\exists ! \zr \in \Pp, \textrm{ such that } \forall \wp \in \Pp,  \wp+\zr=\wp\\
\forall \wp \in \Pp, \exists ! \wp' \in \Pp,  \textrm{ such that }  \wp+\wp'=\zr
}
\end{subequations}
It follows that the additive identity $\zr$ is given by the uniform probability vector. In $\Pp$, it is given by:
\cgather{
\zr = \begin{pmatrix}
1/n & 1/n & \cdots & 1/n 
\end{pmatrix}
}
The ``zero element'' of the group is the uniform distribution.
\subsection{Closed Scalar Multiplication on Probability Vectors}
Since finite dimensional probability vectors reside in $\mathbb{R}^n$, we already have the usual elementwise multiplication by scalars. However, the result of such elementwise scaling will not be a ``probability vector''; the 1-norm will not be unity. Thus, the set $\Pp$ is not closed under the usual multiplication. However we can define a multiplication operation that is indeed closed:
\mltlne{\label{eqmlt}
\forall \alpha \in \mathbb{R}, \wp \in \Pp, \forall i \in \{1,\cdots,n\}, \\ \left ( \alpha\odot\wp\right ) \bigg \vert_i \triangleq \wp_i^\alpha \left ( \sum_j \wp_j^\alpha \right )^{-1} 
}
In the sequel we denote this scalar multiplication by simple concatenation (dropping the $\odot$) if there is no confusion. It is easy to see that:
\begin{subequations}\cgather{
\forall \alpha \in \mathbb{R}, \forall \wp \in \Pp, \alpha \wp \in \Pp \\
\forall \wp \in \Pp, 0\wp = \zr \\
\forall \alpha \in \mathbb{R}, \forall \wp,\wp' \in \Pp, \alpha (\wp+\wp') = \alpha \wp + \alpha \wp'\\
\forall \alpha \in \mathbb{R}, \forall \wp \in \Pp, \alpha\wp + (-\alpha)\wp = \zr\\
\forall \alpha,\alpha' \in \mathbb{R}, \forall \wp \in \Pp, (\alpha\alpha')\wp = \alpha (\alpha' \wp) = \alpha' (\alpha \wp)
}
\end{subequations}
Thus, $\Pp$ has the structure of a \textit{real vector space}, where the group sum is the vector sum, and the above defined product is the scalar product between the vectors and the field elements.

\section{Inner Product on Probability Vectors}\label{secippropbvec}
 The usual ``dot'' product for $n$-dimensional vectors  quite  obviously  applies to elements from $\Pp$. However, this is not the only  consistent inner product on $\Pp$ over the real field.
\begin{defn}[Inner product of  probability vectors]\label{defnip}
We define $\langle \cdot,\cdot \rangle: \Pp[n] \times \Pp[n] \rightarrow \mathbb{R}$ as:
\mltlne{\forall \wp,\wp' \in \Pp[n], \ 
\langle \wp, \wp' \rangle = \sum_{i=1}^{n-1} \ln \left (\wp_i/\wp_{i+1} \right ) \ln \left (\wp'_i/\wp'_{i+1}  \right )
}
\end{defn}
\begin{lem}\label{lemip}
Defn.~\ref{defnip} specifies  an inner product on  $\Pp$, when the latter is considered as a real vector space, with the vector addition and scalar multiplication operations as defined in Eq.~\eqref{eqsum} and Eq.~\eqref{eqmlt} respectively.
\end{lem}
\begin{IEEEproof}
The conditions of Def.~\ref{defii} are easily verified, which completes the proof.
\end{IEEEproof}


\begin{notn} On account of Lemma~\ref{lemip}, we denote the real-valued function introduced in Defn.~\ref{defnip}  as the logarithmic inner product.
\end{notn}
Next, we claim that  $(\Pp,\langle \cdot,\cdot \rangle)$ is infact a \textit{complete} inner product space, $i.e.$ a Hilbert space.  Note that since $\Pp$ only considers probability vectors with non-zero entries, it might seem that we lose completeness: a sequence of such strictly elementwise positive probability vectors can very well converge to one that has zero entries, and hence outside $\Pp$. Nevertheless, we have the following result:

\begin{lem}[Hilbert space of probability vectors]\label{lemcauchy}
$\Pp$ is complete w.r.t. to the norm induced by the logarithmic inner product.
\end{lem}
\begin{IEEEproof} We need to show that every Cauchy sequence in $\Pp$  w.r.t. to the norm induced by the logarithmic inner product converges in $\Pp$.

Let $\{ x_n \}$ be a Cauchy sequence in a normed vector  space $X$, where $d(\cdot,\cdot)$ denotes  the induced metric. We claim:
\cgather{
\forall \epsilon > 0, \exists k \in \mathbb{N}  \textrm{ such that } \forall \ell \geqq k, \ \norm{x_\ell} < \epsilon \tag{Claim A}
}
To establish Claim A, we assume if possible:
\cgather{
  \forall \epsilon > 0, \exists k \in \mathbb{N}  \textrm{ such that } \forall \ell \geqq k, \ \norm{x_\ell} > \epsilon \tag{Assumption A}
}
Now, from definition of Cauchy sequences, we have:
\cgather{
\forall \epsilon > 0, \exists N \in \mathbb{N},  \textrm{ such that } \forall m,n > N, 
\ d(x_m,x_n) < \epsilon
}
Fix some $\epsilon$ and a corresponding $N$. 
Now, for any $x_0 \in X$, and $ \forall m,n > N$, we have:
\calign{
&d(x_m,x_0) \leqq d(x_m,x_n)+d(x_n,x_0) \tag{triangular inequality}\\
\Rightarrow & \epsilon > d(x_m,x_n) \geqq d(x_m,x_0) -d(x_n,x_0) \notag \\
\Rightarrow & d(x_m,x_0) - d(x_n,x_0) \leqq \epsilon
}
Setting $x_0$ to be the vector space zero, we have:
\cgather{\forall m,n > N, \ 
\norm{x_m} - \norm{x_n} \leqq \epsilon \label{eq9}
}
Clearly, if Assumption A holds, we can pick $m,n$ that contradicts Eq.~\eqref{eq9}. Thus, we conclude that the terms of any Cauchy sequence necessarily remains bounded. Since having any zero entry would imply an unbounded induced norm, we conclude that sequences that converge outside $\Pp$ are not Cauchy. It follows that every Cauchy sequence must converge within $\Pp$. This completes the proof.
\end{IEEEproof}
\subsection[Geodesics in the Space of  Prob. Vectors]{Geodesics in the Space of  Probability Vectors}
A geodesic in a metric space is a path connecting two points, such that
no other path has a shorter length.
For completeness, we note here the formal definition of path length, and geodesics.

First, we note the following result:
\begin{lem}\label{lemprobtheta}
Let $\wp_0,\wp_1 \in \Pp$. Then, for $\theta \in [0,1]$,
\mltlne{
\wp_\theta \triangleq \theta \odot \wp_0 \oplus (1-\theta) \odot  \wp_1  = \theta \wp_0 + (1-\theta) \wp_1 \\
\Rightarrow \norm{\wp_{\theta + \delta \theta} - \wp_\theta} = \delta \theta \norm{\wp_0-\wp_1}
} where the norm is  induced by the logarithmic inner product.
\end{lem}
\begin{IEEEproof}
We note that:
\cgather{
\wp_\theta = 
\nlz{\begin{matrix}\cdots & (\wp_0\big\rvert_i )^\theta (\wp_1\big \rvert_i )^{1-\theta }  & \cdots \end{matrix}} \\
\wp_{\theta+\delta\theta} = \nlz{\begin{matrix}\cdots &
(\wp_0\big\rvert_i )^{\theta+\delta\theta} (\wp_1\big \rvert_i )^{1-\theta-\delta \theta } & \cdots \end{matrix}}
\intertext{implying}
\wp_{\theta+\delta\theta} - \wp_\theta = \delta \theta \odot \nlz{ \begin{matrix}\cdots & \wp_0\vert_i \left ( \wp_1 \vert_i\right )^{-1}   & \cdots \end{matrix} } = \delta \theta  (\wp_0-\wp_1)
}
which completes the proof.
\end{IEEEproof}

\begin{figure}
\centering
\tikzexternalenable

\includegraphics[width=3in]{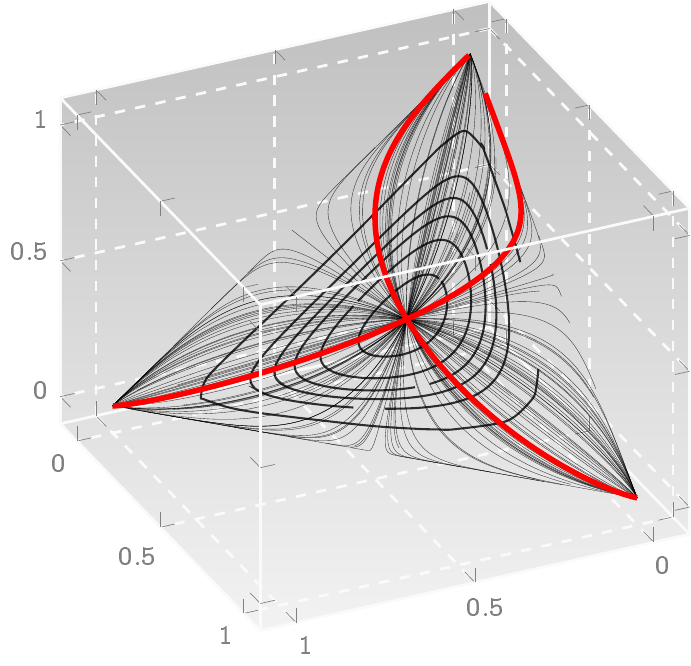}

\captionN{Geodesics on the 2-simplex defined by probability vectors of length 3. The red curves  are two ``straight'' lines perpendicular to each other. The zero of the space is the point defined by the uniform vector $ (1/3,   1/3,  1/3)$, where the orthogonal curves in red intersect.}\label{fig0}
\end{figure} 
\begin{defn}[Length of a Curve and rectifiable curves]
Let $(X, d)$ be a metric space, $I \subset \mathbb{R}$ a non-empty interval, and 
$\gamma: I \rightarrow X$ a Lipshitz-continuous map, $i.e.$, a curve. We define the length $L(\gamma) \in [0,\infty]$:
\cgather{
L(\gamma) \triangleq \sup \sum_{i=1}^n
 d(\gamma(t_{i-1} ), \gamma(t_i ))
}
where the supremum is taken over all $n \in \mathbb{N}$ and all sequences $t_0 \leqq t_1 \leqq \cdots \leqq t_n$
in $I$. We say that $\gamma$ is rectifiable if $L(\gamma) < \infty$.

\end{defn}
Note that length, as defined, is invariant to reparameterization: if $s: I \rightarrow X$ is a curve, and $I' \subset \mathbb{R}$ is another interval, and 
$f:I' \rightarrow I$ is continuous, surjective, and non-decreasing or non-increasing, $i.e.$, $t \leqq t'$ implies $f(t) \leqq f(t')$ or $f(t) \geqq f(t')$, then the curve $s' \triangleq s \circ f: I' \rightarrow X$ satisfies $L(s') = L(s)$.

While the velocity of a curve in an abstract metric space does not make sense, the ``modulus of velocity'', or the metric derivative, is defined as follows:
\cgather{
\abs{\dot{\gamma}(t)} = \limsup_{h\rightarrow 0} \frac{d(\gamma(t+h),\gamma(t))}{h}
}
and using the fact that for almost all $t$, the above limsup is a true limit, we can write:
\cgather{\label{eq18}
L(\gamma) = \int_I \abs{\dot{\gamma}(t)}  dt
}
\begin{defn}[Length Spaces]
For a metric space $(X,d)$, the inner  or length metric associated with
$d$ is the function $d': X \times X \rightarrow [0, \infty]$ defined by
\cgathers{
d'(x,y) \triangleq \inf \left \{ L(\gamma) \vert \gamma \in Lip([0,1],X), \gamma(0) = x, \gamma(1) = y   \right \}
}
where $Lip([0,1],X)$ denotes the set of all Lipshitz continuous maps from $[0,1]$ to $X$. By triangular inequality, we have $d' \geqq d$. If $d'=d$ for all rectifiable curves, then $X$ is a length space.
\end{defn}
\begin{defn}[Geodesic]
In a metric space $(X,d)$, a rectifiable curve $\gamma: I \rightarrow X$ is  geodesic  if $\gamma$ has constant speed and for all $t,t' \in I, t \leqq t'$:
\cgather{\label{eq19}
L(\gamma \vert_{[t,t']}) =d(\gamma(t),\gamma(t'))
} \end{defn}
\begin{rem}\label{remgeodesic}
It follows immediately that a rectifiable curve $\gamma : I \rightarrow X$ is a geodesic if and only if 
\cgather{\label{eq20}
\forall t,t' \in I, \exists \lambda \in (0,\infty), \
d(\gamma(t),\gamma(t')) = \lambda \abs{t-t'}}
\end{rem}
\begin{prop}[Geodesics in $\Pp$]
 For any $\wp_0,\wp_1 \in \Pp$,  the parametric map $\gamma:[0,1] \rightarrow \Pp$  is defined as
\cgather{
\gamma(\theta) = \theta \wp_0 + (1-\theta) \wp_1 
}
\begin{enumerate}
\item $\gamma$ 
is a  geodesic between $\wp_0,\wp_1$.
\item We have the characterization: 
\cgather{\label{eq22}
\norm{\wp_0-\wp_1} = \inf_{\eta} \sqrt{ \int_0^1 \abs{\dot{\eta}(t)}^2 dt} } where  $\eta \in Lip([0,1],\Pp), \eta(0)=\wp_0, \eta(1) = \wp_1$ and 
\item $\gamma$ minimizes the functional on the RHS of Eq.~\eqref{eq22}.
\end{enumerate}
\end{prop}
\begin{IEEEproof}
(1) It follows from Lemma~\ref{lemprobtheta} that $\gamma$ has constant speed equal to $\norm{\wp_0-\wp_1}$, which  immediately verifies Eq.~\eqref{eq20}.
(2) Since $L(\gamma)$ is equal to $\norm{\wp_0-\wp_1}$, we conclude $\Pp$ is  a length space, which then implies the required result from  Eq.~\eqref{eq18}.
(3) By Jensens's inequality~\cite{cover}, 
\cgather{
\int_0^1 \abs{\dot{\eta}(t)} dt \leqq \sqrt{  \int_0^1 \abs{\dot{\eta}(t)}^2 dt  } 
} with equality if and only if $\abs{\dot{\eta}(t)}$ is constant for almost all $t$. Thus, any solution $\eta$ to the functional is necessarily a constant speed geodesic, implying $\gamma$ is a minimizer as required. 
\end{IEEEproof}
\subsubsection{Charting Geodesics in $\Pp$}
We work ou the condition for charting normal curves in $\Pp$. Let two arbitrary curves in $\Pp$ be denoted as:
\cgather{
\gamma(\theta) = \theta \wp_0 + (1-\theta) \wp_1 \\
\eta(\theta) = \theta \wp_0' + (1-\theta) \wp_1'
}
The tangent vectors to these curves at $\theta$ is given by:
\cgather{
\partial \gamma(\theta) = \delta \theta ( \wp_0 - \wp_1) \\
\partial \eta(\theta) = \delta \theta ( \wp_0' - \wp_1') 
}
For the inner product of the tangent vectors to vanish:
\cgather{
( \wp_0 - \wp_1 )\perp (\wp_0'-\wp_1')  
}
If the curves pass through origin, $i.e.$, if $\wp_1=\wp_1'=0$, then the condition 
for the curves to intersect orthogonally at the origin is given by 
$ \wp_0 \perp \wp_0' $. On the other hand, if the curves are orthogonal, and do not pass through the origin, then we can calculate the point of intersection  as:
\calign{
\theta_\star& = \frac{\langle\wp_1-\wp_1',\wp_1-\wp_0 \rangle }{\norm{\wp_1-\wp_0}^2} \\
\wp_\star& = \theta_\star \wp_0 + (1-\theta_\star) \wp_1
}
As a sanity check,  if $\wp_1=\wp_1'=0$, we have $\theta_\star=0$. We map out some of the geodesics for the case of $\vert \Sigma \vert =3 $ in Fig.~\ref{fig0}. 
\begin{rem}\label{remtangentdim}
We note that for the case of a trinary alphabet, the tangent space of $\Pp$ at any point is two dimensional -- the number of vectors mutually orthogonal at any point for such a scenario is $2$.  It is clear that in general, the tangent spaces have dimensionality $\vert \Sigma \vert -1$.
\end{rem}
\section{Modeling Stochastic Processes}

We wish to extend the formalism to stochastic processes. To carry out this extension in a consistent manner, we would require some development. We begin with some notation, and preliminary notions.
\begin{notn}[Sequences over Finite Alpahbet]
\begin{enumerate}
\item 
Let $\Sigma$ be a finite alphabet, and $\Sigma^\omega$ be the set of strictly infinite sequences (or strings) over $\Sigma$ ($\omega$ is not a variable; it is a shorthand for infinite iteration, and this notation is standard in the context of $\omega$-languages~\cite{staiger97}). 
\item The set of finite but unbounded strings over $\Sigma$ is denoted by the Kleene closure of $\Sigma$, namely $\Sigma^\star$~\cite{HMU01}.
\item  For two two sequences $x,y$, the concatenation is written simply as $xy$. 
\item The empty word is denoted as $\lambda$.
\end{enumerate}
\end{notn}

We develop a slightly non-standard formalism of modeling stochastic processes, compared to what is generally encountered in the literature. We are interested in processes that take values in a finite set (the specified alphabet), instead of the real line, and our intentional departure from the standard formalism underscores the  connection to formal languages arising from the finite valued nature of such processes.
\begin{defn}[Cantor Topology on $\Omega$-Languages] Let $\B=\{x\Sigma^\omega: x \in \Sigma^\star\}$ be a family of sets of infinite sequences. Note $x\Sigma^\omega$ denotes the set of all strictly infinite sequences which have $x$ as the common prefix. It is easy to check that $\B$ qualifies as a basis for inducing a topology. In particular, we have:
\begin{enumerate}
\item $\bigcup \B = \Sigma^\star \Sigma^\omega = \Sigma^\omega$
\item $\forall B_1,B_2 \in \B \Rightarrow B_1 \cap B_2 = \varnothing $ or $ B_1 \subseteqq B_2$ or $B_2 \subseteqq B_1$, which guarantees that $\forall z \in B_1 \cap B_2 \Rightarrow \exists B \in \B$ such that $ z \in B_1\cap B_2\cap B$.
\end{enumerate}
It follows that exists an unique topology for which $\B$ is a base. We denote this topology as $\U$. Indeed, this is the Cantor topology induced by the Tychonoff construction~\cite{munkres2000topology} on  countable product of finite discrete sets~\cite{staiger97} (in this case this finite  set is the alphabet).
\end{defn}
We note that on account of $\B$ being the base for $\U$, every open set in $\U$ may be written as a union of elements of $\B$. Since $\B$ is countable, it follows that every open set is of the form $L\Sigma^\omega, L \subseteqq \Sigma^\star$.
\begin{defn}[Borel $\sigma$-algebra $\F$]
$\F$ is defined as the smallest $\sigma$-algebra containing $\U$, implying that $\F$ is the Borel $\sigma$-algebra wrt $\U$. It trivially follows that, every measurable set is also of the form 
$L\Sigma^\omega, L \subseteqq \Sigma^\star$.
\end{defn}

Using $\F$, we can now define a probability space $(\Sigma^\omega,\F,\mu)$, which models a stochastic process, assigning probabilities to sets of strictly infinite sample paths. Note, in particular, that a strictly infinite single sample path is not measurable (such sets are not included in $\F$); only sets that are of the form specified before, are; and after a finite length include all possible extensions into future.

We also consider here the map $T: \Sigma^\omega \rightarrow \Sigma^\omega$ defined by:
\cgather{
T(x_1x_2x_3\cdots) = x_2x_3\cdots
}
It is immediate that $T$ is measurable wrt $\F$.
In going forward, we assume:
\cgather{
\forall A \in \F, \mu(T^{-1}A) = \mu(A) \tag{Stationarity}
}
imposing that we are considering only stationary processes. Additionally, we assume:
\cgather{
\forall A \in \F, \mu(A \Delta T^{-1}A ) = 0 \Rightarrow \mu(A) \in \{0,1\} \tag{Ergodicity}
}
which ensures that our systems of interest are also ergodic.
\begin{rem}[Relationship to Standard Formalism]\label{remstd}
There is a quite obvious connection to the  standard formalism. Namely, the finite dimensional distributions can be identified as:
\cgather{
Pr(X_1X_2, \cdots, X_n=x_1x_2\cdots x_n) = \mu(x_1x_2\cdots x_n \Sigma^\omega)
}
Noting that:
\mltlne{
\sum_{x_n \in \Sigma} \mu(x_1x_2\cdots x_n \Sigma^\omega) \\= \mu\left (\displaystyle \bigcup_{x_n \in \Sigma}x_1x_2\cdots x_n \Sigma^\omega\right ) = \mu(x_1x_2 \cdots x_{n-1} \Sigma^\omega)
}
implies that the finite dimensional distributions are Kolmogorov consistent, and hence using Kolmogorov Extension theorem~\cite{doob1953stochastic,kolmogorov1950foundations}, we can go back and forth between the two formalisms.
\end{rem}
\subsection{States and Transition Structure}
\begin{defn}[Probabilistic Nerode Equivalence \& Causal States]\label{defnerode} We define an  relation on the set of all finite but unbounded strings, $i.e.$ the set $\Sigma^\star$, as follows:
\cgathers{
\forall \omega_1,\omega_2 \in \Sigma^\star, 
\omega_1 \simn \omega_2, \textrm{if } \left \{ \begin{array}{ll}
\mu(\omega_1\Sigma^\omega) = \mu(\omega_2\Sigma^\omega) = 0\\
\textrm{or} \\
\mu(\omega_1\Sigma^\omega) \neq 0, \textrm{ and}\\ \mu(\omega_2\Sigma^\omega) \neq 0, \textrm{ and}\\
\forall z \in \Sigma^\star, 
\frac{\mu(\omega_1z \Sigma^\omega)}{\mu(\omega_1 \Sigma^\omega)} = \frac{\mu(\omega_2z \Sigma^\omega)}{\mu(\omega_2 \Sigma^\omega)}
\end{array}\right.
}
It is easy to see that this is actually a right invariant equivalence relation, $i.e.$,
\cgather{
x \simn y \Rightarrow \forall z \in \Sigma^\star, xz \simn yz
}
and hence intuits the notion of states. We define the ``causal states'' of the process, as the equivalence classes of this relation.
\end{defn}
\begin{defn}[Symbolic Derivative]
For $x \in \Sigma^\star$, with $\mu(x\Sigma^\omega) > 0$, the symbolic derivative $\phi: \Sigma^\star \rightarrow \Pp[{{\abs{\Sigma}}}]$  is a probability distribution over the alphabet, defined as:
\cgather{\label{eq35}
\phi(x) \big \rvert_\sigma = \frac{\mu(x\sigma\Sigma^\omega: x \in \Sigma^\star, \sigma \in \Sigma)}{\mu(x\Sigma^\omega: x \in \Sigma^\star)} 
}
Clearly, we have for any $x \in \Sigma^\star$, with $\mu(x\Sigma^\omega) > 0$, $\sum_{\sigma \in \Sigma} \phi(x) \vert_\sigma = 1$. We refer to  $\phi(x)$ as the symbolic derivative at $x$, and denote it as $\phi_x$.
\end{defn}
It is clear that for strings $x,x' \in \Sigma^\star$, we have:
\cgather{
x \simn x' \Leftrightarrow \forall y \in \Sigma^\star, \phi_{xy} = \phi_{x'y}
}
\begin{lem}[Sufficiency of Symbolic Derivatives]\label{lemsuff} 
The set of  symbolic derivatives at all finite strings, $i.e.$,   $\{\phi_x: x \in \Sigma^\star\}$ uniquely specifies a  measure $\mu$ on the measurable space $(\Sigma^\omega, \F)$.
\end{lem}
\begin{IEEEproof}
$\mu$ is uniquely specified by the  recursions:
\begin{subequations}\cgather{
\forall \sigma \in \Sigma, \mu(\sigma\Sigma^\omega) = \phi_\lambda\big \vert_\sigma\\
\forall x \in \Sigma^\star, \sigma \in \Sigma, \mu(x\sigma\Sigma^\omega) = \left \{ \begin{array}{ll}\mu(x\Sigma^\omega) \phi_x \big \vert_\sigma & \textrm{if } \mu(x\Sigma^\omega) > 0\\0 & \textrm{otherwise}\end{array}\right.
}
\end{subequations}
This completes the proof.
\end{IEEEproof} 
\begin{rem}
Another approach to proving the claim in Lemma~\ref{lemsuff} would be to show that the complete set of symbolic derivatives induces a complete set of finite dimensional distributions (FDD) via:
\calign{\label{eq38}
Pr(X_1) &= \phi_\lambda\\
 Pr(X_1\cdots X_nX_{n+1} &= x_1\cdots x_n \sigma )\notag \\ &= Pr(X_1\cdots X_n= x_1\cdots x_n)\phi_x \big \vert_\sigma
}
which are clearly Kolmogorov consistent, and hence via the Kolmogorov extension theorem~\cite{kolmogorov1950foundations} induces a stochastic process, which is FDD-equivalent to $(\Sigma^\omega, \F, \mu ) $ (See Remark~\ref{remstd}).
\end{rem}
\subsubsection{States As A Random Variable}
We do not wish to identify any initial state of our processes of interest. Thus, given an observed sequence, we assume that arbitrary sequences could have transpired prior to the observations. This induces the notion of a causal state as a random variable:
\cgather{
\eqcl: (\Sigma^\omega, \F, \mu) \rightarrow (Q, \F[Q],Pr)
}
where $Q$ is the set of equivalence classes (the atmost countable state space), $\F[Q]$ is an appropriate $\sigma$-algebra (which generally we will take to be the power set of $Q$), and $Pr$ is the pushforward of the measure $\mu$. Thus, we have:
\calign{\forall q \in Q, Pr(q) &= \mu(\eqcl^{-1}(q)) \notag \\ &= \mu\left ( x\Sigma^\omega: x \in \Sigma^\star \wedge \eqcl[x] = q \right )
}
\begin{defn}[Conditioning on Observations]\label{defobs}
Given some observed sequence $x_0 \in \Sigma^\star$, we condition as follows:
\calign{
Pr(q \vert x_0) & \triangleq \mathsf{const.} \times \mu(\eqcl^{-1}(q) \vert \Sigma^\star  x_0 ) 
}
\end{defn}
%
\begin{lem} Assuming stationarity,
\calign{
 \forall x_0 \in \Sigma^\star, & \mu(x_0 \Sigma^\omega) > 0  \Rightarrow \notag    \\ 1) \mspace{20mu} & Pr(q \vert x_0)  =  \frac{\mu \left ( yx_0 \Sigma^\omega : y \in \Sigma^\star \wedge \eqcl[yx_0] = q   \right )}{\mu( x_0 \Sigma^\omega) } \\  2)  \mspace{20mu} &   \sum_{q \in Q} Pr(q \vert x_0) = 1}
\end{lem}
\begin{IEEEproof} Denoting the normalizing constant as $C$,
\cgather{
Pr(q \vert x_0)  =  C \mu \left ( yx_0 \Sigma^\omega : y \in \Sigma^\star \wedge \eqcl[yx_0] = q   \right )
}
which implies (invoking stationarity in the last step)
\cgather{
C^{-1} =  \mu \left ( yx_0 \Sigma^\omega : y \in \Sigma^\star   \right ) =  \mu \left ( \Sigma^\star x_0 \Sigma^\omega \right )
 = \mu(x_0\Sigma^\omega)}
The second statement is immediate.\end{IEEEproof}
\begin{rem}
In Definition~\ref{defobs} we assume that an observed sequence is the suffix of the complete transpired sequence; any finite  sequence of values could have occurred before the specific observations. Also, note:
\cgather{
\forall q \in Q, Pr(q\vert \lambda) = Pr(q)
}
\end{rem}
\subsubsection[Probabilistic Automata]{Probabilistic Automata Generators}
\begin{defn}[Probabilistic Automata (PA)]\label{defpa}
A probabilistic automata is a 4-tuple $(\Sigma,Q, \delta,\pitilde)$, where $\Sigma$ is a finite set (the alphabet),  $Q \subseteqq \mathbb{N}$ is the state space, $\delta: Q \times \Sigma \rightarrow Q$ is the transition map,  and $\pitilde : Q \times \Sigma \rightarrow [0,1]$ specifies the state-specific transition probabilities, satisfying $\forall q \in Q, \sum_{\sigma \in \Sigma} \pitilde(q,\sigma) = 1$. 
\end{defn}
\begin{defn}
We use the following terminology:
\calign{
\Pitilde_{ij} &\triangleq \pitilde(q_i,\sigma_j) \tag{Morph Matrix}\\
\Pi_{ij} &\triangleq \sum_{\mathclap{\sigma: \delta(q_i,\sigma) = q_j}}\pitilde(q_i,\sigma)  \tag{Transition Probability Matrix}\\
\Gamma_\sigma \big \rvert_{ij} &\triangleq \left \{ \begin{array}{ll} \pitilde(q_i,\sigma) & \textrm{if } \delta(q_i,\sigma) = q_j\\
0 & \textrm{otherwise} \end{array}\right. \tag{Event-specific Transition Matrix}\\
\textrm{Note that,} & \textrm{ we have } 
\sum_{\sigma \in \Sigma} \Gamma_\sigma = \Pi \notag
}
 We say a probabilistic automata $G=(\Sigma,Q,\delta,\pitilde)$  is a probabilistic finite state automata (PFSA) if $\abs{Q} < \infty$. In that case, we have the morph, transition probability, and the event specific transition probability matrices as respectively of dimensions $\abs{Q} \times \abs{\Sigma}, \abs{Q} \times \abs{Q}, \abs{Q}\times \abs{Q}$.
\end{defn}
Probabilistic automata are convenient representations for stationary ergodic finite-valued stochastic processes. We say that an automaton encodes a process if all finite dimensional distributions (FDD) may be recovered from it, $i.e.$,   the model represents the  process upto FDD equivalence. 
\begin{lem}[Probabilistic Automata to Stochastic Process]\label{lempatoms}
$\pax$ induces a stationary stochastic process if 
\mltlne{
\exists \wp' \in [0,1]^{\abs{Q}}, \textrm{ with }  \sum_j \wp'_j = 1, \\
\textrm{s.t. } \forall i \in Q,  \sum_{j\in Q} \wp'_j \Pi_{ij} = \wp_i
}
\end{lem}
\begin{IEEEproof}
We define  $\wp'_x, \phi'_x,  x \in \Sigma^\star$ as follows:
\begin{subequations}
\calign{
&\wp'_\lambda = \wp'\\
&\wp'_{x\sigma} = \nlz{\wp'_x \Gamma_\sigma}\\
&\phi'_{x} = \wp'_x \Pitilde
}
\end{subequations}
We then construct a set of Kolmogorov consistent set of finite dimensional distributions recursively as:
\calign{\label{eq38}
Pr(X_1) &= \phi'_\lambda\\
 Pr(X_1\cdots X_nX_{n+1} &= x_1\cdots x_n \sigma )\notag \\ &= Pr(X_1\cdots X_n= x_1\cdots x_n)\phi'_x \big \vert_\sigma
\label{eq52}}
which, then via invocation of the Kolmogorov Extension Theorem~\cite{kolmogorov1950foundations} induces a FDD equivalent measure space $(\Sigma^\omega, \F, \mu)$. The recursive construction of the finite dimensional distributions in Eqns.~\eqref{eq38},\eqref{eq52} have no dependence on time shifts, and hence guarantee stationarity. This completes the proof.
\end{IEEEproof}
We use the following notation:
\begin{notn}
If  $(\Sigma,Q,\delta,\pitilde)$ encodes in the sense of Lemma~\ref{lempatoms} the stationary stochastic process  arising from $(\Sigma^\omega,\F,\mu)$ then we write:
\cgather{
(\Sigma,Q,\delta,\pitilde) \models  (\Sigma^\omega,\F,\mu)
}
\end{notn}
The importance of probabilistic automata based encodings arises from the following proposition.
\begin{prop}[Existence of Canonical Encoders]\label{propencode}
For every stationary ergodic process generated by the measure space $(\Sigma^\omega, \F, \mu)$, we have a $\pax$, such that:
\cgather{
\pax \models (\Sigma^\omega, \F, \mu)
}
\end{prop}
\begin{IEEEproof}
%
 A stationary ergodic process arising from the triple $(\Sigma^\omega,\F,\mu)$ induces a  $(\Sigma,Q,\delta,\pitilde)$ as follows (this construction is referred to in the sequel as the \underline{canonical encoding}):
\begin{enumerate}
\item Identify $Q$ as the set of equivalence classes for $\simn$. 
\item Identify the transition structure  as:
\cgathers{
\forall q \in Q, \textrm{ choose } x \in \Sigma^\star, \textrm{ s.t. } [x] = q. \textrm{ Then } \forall \sigma \in \Sigma, \\  \delta([x],\sigma) = [x \sigma], 
\pitilde([x],\sigma) = \phi_x \big \vert_\sigma
}
\end{enumerate}
We claim that the symbolic derivatives are recoverable from  $(\Sigma,Q,\delta,\pitilde)$. 
To establish this claim, we will construct a set of recursive relationships that would allow us to recover the complete set of symbolic derivatives. We denote $\wp_x \rvert_i \triangleq Pr(q_i \vert x)$, and proceed by noting:
\calign{
\sum_{j \in Q} \wp_\lambda\vert_j \Pi_{ji} &= \sum_{j\in Q} \mu(x\Sigma^\omega: \eqcl[x]=j) \frac{\mu(x\sigma\Sigma^\omega:\eqcl[x\sigma]=i \wedge \eqcl[x]=j)}{\mu(x\Sigma^\omega:\eqcl[x]=j)} \notag  \intertext{(where we assume $\mu(x\Sigma^\omega) > 0$)}
&=\sum_{j\in Q} \mu(x\sigma\Sigma^\omega:\eqcl[x\sigma]=i \wedge \eqcl[x]=j) \\
&=\mu(y\Sigma^\omega:\eqcl[y]=i) = \wp_\lambda\vert_i \label{eqcond1}
}
which implies that a unique stationary distribution corresponding to $\Pi$ exists, which is given by $\wp_\lambda$. Next, we observe:
\calign{
&\wp_{x\sigma} \vert_i = \frac{\mu(yx\sigma\Sigma^\omega: [yx\sigma]=i)}{\mu(x\sigma\Sigma^\omega)} \tag{from Definition~\ref{defobs}} \intertext{(Assuming $\mu(x\sigma\Sigma^\omega) > 0$ and $\mu(x\Sigma^\omega) > 0$)} &=\frac{\sum_{j\in Q}\mu(yx\sigma\Sigma^\omega: \eqcl[yx]=j \wedge \eqcl[yx\sigma]=i)}{\mu(x\sigma\Sigma^\omega)} \\
&=\sum_{j\in Q} \frac{\mu(z\Sigma^\omega: \eqcl[z]=j)}{\mu(x\Sigma^\omega)} \frac{\mu(yx\sigma\Sigma^\omega: \eqcl[yx]=j \wedge \eqcl[yx\sigma]=i)}{\mu(z\Sigma^\omega: \eqcl[z]=j)}\notag  \\ &
\mspace{340mu} \vartimes  \frac{\mu(x\Sigma^\omega)}{\mu(x\sigma\Sigma^\omega)} \\
& =\sum_{j\in Q} \frac{\mu(yx\Sigma^\omega: \eqcl[yx]=j)}{\mu(x\Sigma^\omega)} \frac{\mu(yx\sigma\Sigma^\omega: \eqcl[yx]=j \wedge \eqcl[yx\sigma]=i)}{\mu(yx\Sigma^\omega: \eqcl[yx]=j)}  \notag\\ &
\mspace{340mu}\vartimes\frac{\mu(x\Sigma^\omega)}{\mu(x\sigma\Sigma^\omega)} \\
&=\nlz{\sum_{j\in Q} \frac{\mu(yx\Sigma^\omega: \eqcl[yx]=j)}{\mu(x\Sigma^\omega)} \frac{\mu(yx\sigma\Sigma^\omega: \eqcl[yx]=j \wedge \eqcl[yx\sigma]=i)}{\mu(yx\Sigma^\omega: \eqcl[yx]=j)}} \notag\\
&=\nlz{\sum_{j\in Q} \frac{\mu(yx\Sigma^\omega: \eqcl[yx]=j)}{\mu(x\Sigma^\omega)} \frac{\mu(z\sigma\Sigma^\omega: \eqcl[z]=j \wedge \eqcl[z\sigma]=i)}{\mu(z\Sigma^\omega: \eqcl[z]=j)}} \notag\\ &= \nlz{\sum_{j \in Q}\wp_x\vert_j \Gamma_\sigma \vert_{ji}} \label{eqcond2}
}
Finally, we note:
\cgather{
\shoveleft \sum_{q_j\in Q} \wp_x \vert_{q_j} \pitilde(q_j,\sigma) = \sum_{q_j\in Q} Pr(q_j \vert x) \pitilde(q_j,\sigma)\intertext{(Assuming $\mu(x\sigma\Sigma^\omega) > 0$)}
=\sum_{j \in Q} \frac{\mu(yx\Sigma^\omega: \eqcl[yx]=j)}{\mu(x\Sigma^\omega)} \frac{\mu(yx\sigma\Sigma^\omega:\eqcl[yx]=j)}{\mu(yx\Sigma^\omega: \eqcl[yx]=j)}\\ 
=\sum_{j \in Q} \frac{\mu(yx\sigma\Sigma^\omega:\eqcl[yx]=j)} {\mu(x\Sigma^\omega)}\\ = \frac{\mu(\Sigma^\star x\sigma\Sigma^\omega)} {\mu(x\Sigma^\omega)} = \frac{\mu(x\sigma\Sigma^\omega)} {\mu(x\Sigma^\omega)} = \phi_x\vert_\sigma\label{eqcond3}
}
where stationarity is  invoked in Eq.~\eqref{eqcond3}. We note that Eqns.~\eqref{eqcond1},\eqref{eqcond2}, and \eqref{eqcond3}, may be summarized as (representing $\wp_x$ as a row vector to use matrix notation):
\begin{subequations}
\calign{
\wp_\lambda \Pi &= \wp_\lambda \intertext{And, $
\forall x \in \Sigma^\star,  \sigma\in \Sigma, \textrm{ s.t. } \mu(x\Sigma^\omega) >0,\mu(x\sigma\Sigma^\omega) >0$,} \wp_{x\sigma} &= \nlz{ \wp_x \Gamma_\sigma}\\
\phi_x &= \wp_x \Pitilde 
}
\end{subequations}
which gives us the desired recursions that recover the complete set of symbolic derivatives $\{\phi_x: x \in \Sigma^\star, \mu(x\Sigma^\omega) > 0\}$. 
Lemma~\ref{lemsuff} then guarantees that the measure $\mu$ may be constructed  from $(\Sigma,Q,\delta,\pitilde)$.
\end{IEEEproof}
\begin{notn}
The canonical encoding described in Proposition~\ref{propencode} is denoted as $\pan$.
\end{notn}
%
\begin{rem}
Finiteness of the state space is not invoked in proving the existence of PA encoders in Proposition~\ref{propencode}, and hence $Q$ in the construction is atmost countable.
\end{rem}
\begin{defn}[Closed Restriction]\label{defclosedrestruction}
A closed restriction of $\pax$ is a  model $\pax[']$ such that:
\begin{subequations}
\cgather{
\varnothing \neq Q' \subseteqq Q\\
\forall \sigma \in \Sigma, q'  \in Q', \delta'(q',\sigma) \in Q' \\
\forall \sigma \in \Sigma, q'  \in Q', \pitilde'(q',\sigma) = \pitilde(q',\sigma)
}
\end{subequations}
The set of all closed restrictions of a probabilistic automaton $G=\pax$ is denoted as $\clx{G}$.
A closed restriction $H \in \clx{G}$ is a minimal closed restriction if 
\cgather{
\clx{H} = \{ H \} 
}
The set of all minimal closed restrictions of a probabilistic automaton $G$ is denoted as $\cly{G}$. Note that we have 
\cgather{
\cly{G} \subseteqq \clx{G}
}
\end{defn}
\begin{defn}[Probability of Closed Restriction]\label{defcrprob}
For a closed restriction $H \in \clx{G}$, and $\pax \models \wax$,  the total probability $Pr(H)$ is defined as follows:
\cgather{
\textrm{If } H = \pax['], Pr(H) \triangleq \sum_{q \in Q'} \mu(x\Sigma^\omega : \eqcl[x]=q)
}
\end{defn}

\begin{lem}[Closed Restriction]\label{lemclosed}
If $\wax$ is stationary, ergodic with $\pax \models \wax$ (without loss of generality according to Proposition~\ref{propencode}), then:
\cgather{
\exists ! H \in \cly{\pax}, \textrm{ s.t. }  Pr(H)=1
}
\end{lem}
\begin{IEEEproof}
Indexing elements of $\cly{\pax}$ as $G_i= (\Sigma, Q_i, \delta^i, \pitilde^i)$, it follows immediately:
\cgather{
G_i \neq G_j \Rightarrow Q_i \cap Q_j = \varnothing
}
Recalling that $\forall i, Q_i \subseteqq Q$, let us define:
\cgather{
L_i \triangleq \bigcup_{q\in Q_i} \big \{x: x \in \Sigma^\star, \eqcl[x]=q  \big \}
\intertext{and we conclude:}
G_i \neq G_j \Rightarrow L_i \cap L_j = \varnothing
}
Since, $\forall i, G_i$ are minimal closed restrictions, we have (considering the standard shift map $T$):
\cgather{
T^{-1} (L_i \Sigma^\omega) = L_i \Sigma^\omega
}
and then ergodicity of $\pax$ implies:
\cgather{
\forall i, \mu(L_i\Sigma^\omega) \in \{0,1\}
}
Finally, $\bigcup_i L_i \subseteqq \Sigma^\star$, implies that there 
exists a unique minimal closed restriction with full measure, completing the proof.
\end{IEEEproof}
\begin{notn}[Unique Minimal Closed Restriction]\label{notclosed}
If $\wax$ is stationary, ergodic with $\pax \models \wax$, the unique minimal closed restriction $H \in \cly{\pax}$ with $Pr(H) = 1$ is denoted as $\CLX{\pax}$. Note if we denote $\paz[']=\CLX{\pax}$, then  $\delta', \pitilde'$ in $\paz[']$ are appropriate  restrictions of the corresponding functions in $\pax$ to $Q'$.
\end{notn}
We show next that the unique minimal closed restriction is sufficient to model the process, and consists of all the non-trivial states in the original model.
\begin{lem}[Sufficiency of Minimal Closed Restriction]\label{lemsuffun}
If $\wax$ is stationary, ergodic with $\pax \models \wax$, the unique minimal closed restriction $\CLX{\paz}=(\Sigma,Q^\star,\delta^\star,\pitilde^\star)$ satisfies:
\begin{subequations}
\cgather{
\forall q \in Q^\star, Pr(q) > 0 \label{eq78a}\\
\forall q \in Q, Pr(q) > 0 \Rightarrow q \in Q^\star  \label{eq78b}\\
\CLX{\pax} \models \wax \label{eq78c}
}
\end{subequations}
\end{lem}
\begin{IEEEproof}
Let if possible we have a state $q_0$ such that:
\cgather{
Pr(q_0) =0 \wedge  q _0 \in Q^\star
\intertext{
Then, recalling that $q_0$ is also a state in $\pax$, we have:}
\mu(x\Sigma^\omega: x \in \Sigma^\star \wedge \eqcl[x]=q_0) = 0
}
Since,
\cgather{
\mu(x\Sigma^\omega) = 0 \Rightarrow \forall y \in \Sigma^\star, \mu(xy\Sigma^\omega) = 0 \intertext{it follows}
\eqcl[x']=q_0 \Rightarrow  \forall y \in \Sigma^\star, \eqcl[x'y]=q_0
}
which then implies that $(\Sigma,\{q_0\},\delta',\pitilde')$, with $\delta', \pitilde'$ appropriate restrictions of $\delta,\pitilde$,  defines a minimal closed restriction (contradiction). This establishes Eq.~\eqref{eq78a}. Eq.~\eqref{eq78b} follows immediately from  $Pr(\paw)=1$ (Lemma~\ref{lemclosed}).

To establish Eq.~\eqref{eq78c}, 
we note that if the stationary probability vector for $\pax$ is denoted as $\wp$ (which exists on account of Lemma~\ref{lemclosed} and Notation~\ref{notclosed}), then 
a stationary probability vector $\wp^\star$ exists for $\paz[\star]$, and is given simply as the restriction:
\cgather{
\wp^\star_x = \wp_x \big\rvert_{Q^\star} 
}
Also, note that  Eqns.~\eqref{eq78a},\eqref{eq78b} establish that $\wp^\star$ accounts for all non-zero entries in $\wp$.
Now, following the construction in Lemma~\ref{lempatoms}, we define:
\calign{
&\wp_\lambda = \wp \\
&\wp_{x\sigma} = \nlz{\wp_x \Gamma_\sigma}\\
&\phi'_{x} = \wp'_x \Pitilde 
}
and for the case of $\paw$, 
\calign{
&\wp^\star_\lambda = \wp^\star \\
&\wp^\star_{x\sigma} = \nlz{\wp^\star_x \Gamma^\star_\sigma}\\
&\phi^\star_{x} = \wp^\star_x \Pitilde^\star}
where $\Gamma^\star_\sigma, \Pitilde^\star$ are the corresponding Event-specific Transition matrix, and the morph  matrix (See Definition~\ref{defpa}) for $\paz[\star]$. We claim that:
\cgather{
\forall x \in \Sigma^\star,  \phi_x = \phi^\star_x
}
which follows immediately from noting that since $\paz[\star]$ is a minimal closed restriction, no transition from any state in $Q^\star$  by any $\sigma \in \Sigma$ takes us outside the set $Q^\star$, implying that since $\wp_\lambda  \big \vert_{Q\setminus Q^\star}$ is a zero vector, $\forall x \in \Sigma^\star, \wp_x \big \vert_{Q\setminus Q^\star}$ is also a  zero vector. Hence, the contribution from states outside $Q^\star$ to $\phi_x$ is zero for all $x$. Thus, the measure specified on $(\Sigma^\omega, \F)$ by $\paz[\star]$ coincides with that induced by $\pax$ (Lemma~\ref{lemsuff}). This completes the proof.
\end{IEEEproof}
To paraphrase Lemma~\ref{lemsuffun}, given any probabilistic automata that models a finite values stationary ergodic process, the unique minimal closed restriction also models the process. And this result holds for atmost countable state spaces.
We next establish that the unique minimal closed restriction of the canonical model constructed in Proposition~\ref{propencode} is infact an unique minimal realization of the process.
\begin{prop}[Existence of Minimal Models]\label{propmin}
If  an arbitrary probabilistic automata $\pax \models \wax$, then:
\begin{enumerate}
\item $\paz[\star]=\CLX{\pax}$ induces an equivalence relation $\sim'$ on $\Sigma^\star$, where there is a one-to-one mapping from the equivalence classes of $\sim'$ to $Q^\star$.
\item $\sim'$ is a refinement of $\simn$.
\end{enumerate}
\end{prop}
\begin{IEEEproof}
Since $\pax \models \wax$, denoting:
\cgather{
\wp_x \vert_i = \frac{\mu(yx\Sigma^\omega : y \in \Sigma^\star \wedge \eqcl[yx]=i)}{\mu(x\Sigma^\omega)}
}
we can define an equivalence on $\Sigma^\star$ as follows:
\cgather{
\forall x,y \in \Sigma^\star, x \sim' y \textrm{ if } \forall z \in \Sigma^\star, \wp_{xz} = \wp_{yz}
}
We note that there exists a one-to-one map $\zeta$ from $Q^\star$ to the equivalence classes of $\sim'$:
\cgather{
\forall i \in Q^\star, \zeta(i) = \eqcl[x]', \forall x \textrm{ s.t. } \wp_x\vert_i=1
}
This establishes Statement (1).
For Statement (2), we note:
\cgather{
x \sim' y \Rightarrow \forall z \in \Sigma^\star, \wp_{xz} = \wp_{yz} \\ \Rightarrow \forall z \in \Sigma^\star, \phi_{xz} = \wp_{xz} \Pitilde = \wp_{yz}\Pitilde = \phi_{yz}
}
which completes the proof.
\end{IEEEproof}
Thus, it follows that unique minimal closed restriction of the canonical encoding, 
whose states correspond to the non-trivial (consisting of non-zero probability strings) equivalence classes of $\simn$, represent the unique minimal model, in the sense of representing the coarsest equivalence on $\Sigma^\star$.
For probabilistic finite state automata encoders, we have the following result on the state space sizes.
\begin{cor}[To Proposition~\ref{propmin}: Minimal Models in Finite State Space Case]\label{corpfsamin}
Let an arbitrary $\pax \models \wax$, and  $\pax[']=\CLX{\pan}$ be the  unique minimal closed restriction of the canonical encoding. If  $\abs{Q} < \infty$, we have: 
\begin{subequations}
\cgather{
\abs{Q^\circ} < \infty\\
\abs{Q} \geqq \abs{Q'}\\
\abs{Q} = \abs{Q'} \Rightarrow \sim' \equiv \simn
}
\end{subequations}
\end{cor}
\begin{IEEEproof} Denote $\paz[\star]=\CLX{\pax}$.

Statement (1): 
Since $\abs{Q} < \infty$, it follows from the definition of closed restrictions that
$\abs{Q^\star}< \infty$. It then follows from Proposition~\ref{propmin} and the definition of canonical encodings that 
$\abs{Q^\circ} < \infty$, as required.

Statement (2): It follows from Proposition~\ref{propmin}:
\cgather{
\abs{Q} \geqq \abs{Q^\star} \geqq \abs{Q^\circ} \geqq \abs{Q'}
}
Statement (3): Follows immediately from Proposition~\ref{propmin}.
\end{IEEEproof}
Thus, the unique minimal closed restriction of the canonical encoding $\CLX{\pan}$ is the minimal model unique upto a renaming of the states.
\begin{rem}[Minimal and Non-minimal Realizations of Models]
While the minimal realization is unique, it is trivial to generate non-minimal realizations of encoders. In particular, any refinement of the $\simn$-equivalence gives us a non-minimal probabilistic automata correctly  encoding the same process.
\end{rem}
\begin{rem}
Corollary~\ref{corpfsamin} uses finiteness of the state spaces; the preceding results hold for atmost countable states.
\end{rem}
\subsection{Synchronization}
\begin{figure}
\centering   

\includegraphics[width=3.5in]{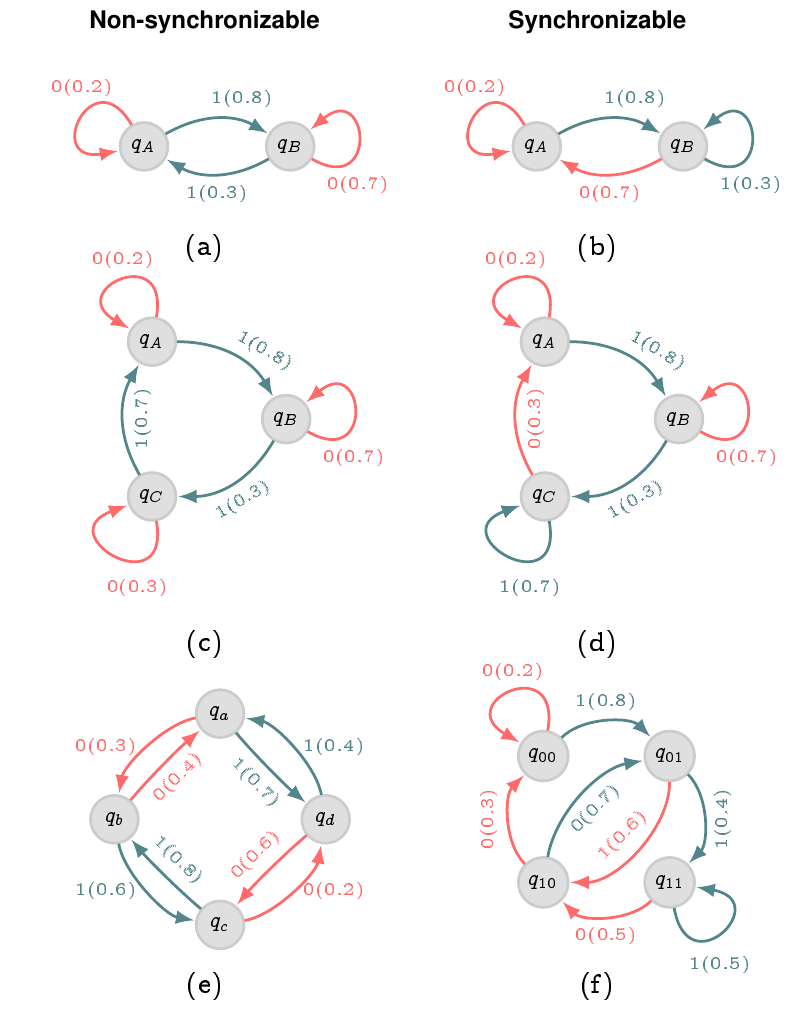}
\captionN{Synchronizable \& non-synchronizable Models. Proposition~\ref{propeps} establishes that non-synchronizable models are still $\epsilon$-synchonizable.}\label{figsync}
\end{figure}
In the sequel, unless otherwise mentioned, we always consider the unique minimal closed restriction of the canonical embedding by  $\pax \models \wax$, where $\wax$ is always assumed to be stationary, ergodic. We do not assume finiteness of the state spaces, unless mentioned explicitly.
\begin{lem}[Balance Lemma]\label{lembalance}
For $\pax \models \wax$, given some state probability vector  $p$, where as usual $ \forall i,  p_i > 0, \sum_i p_i =1$,  we have:
\mltlne{
\exists \sigma_\star \in \Sigma, \frac{\pitilde(i,\sigma_\star)}{\sum_ip_i\pitilde(i,\sigma_\star)} < 1 \Leftrightarrow \exists \sigma' \in \Sigma,  \frac{\pitilde(i,\sigma')}{\sum_ip_i\pitilde(i,\sigma')} > 1 \notag
}
\end{lem}
\begin{IEEEproof}
Let us assume for some $\sigma_\star \in \Sigma$, 
\cgather{
\frac{\pitilde(i,\sigma_\star)}{\sum_ip_i\pitilde(i,\sigma_\star)} < 1 
}
Then, either the claim from left to right is true, or we have for all but  some $\sigma' \in \Sigma$:
\cgather{
\forall \sigma \in \Sigma\setminus \{\sigma',\sigma_\star\}, \frac{\pitilde(i,\sigma)}{\sum_ip_i\pitilde(i,\sigma)} \leqq 1
}
But, then for $\sigma'$, we have:
\mltlne{
\frac{\pitilde(i,\sigma')}{\sum_ip_i\pitilde(i,\sigma')} = \frac{1-\sum_\sigma \pitilde(i,\sigma)}{1- \sum_\sigma\sum_ip_i\pitilde(i,\sigma)} \mspace{80mu} \\ \geqq \frac{1- \pitilde(i,\sigma_\star)   - \sum_{\sigma \in \Sigma \setminus\{\sigma',\sigma_\star\}} \sum_ip_i\pitilde(i,\sigma)}{1- \sum_{\sigma \in \Sigma \setminus\{\sigma'\}}\sum_ip_i\pitilde(i,\sigma)} \\ >
\frac{1-\sum_\sigma \sum_ip_i\pitilde(i,\sigma)}{1- \sum_\sigma\sum_ip_i\pitilde(i,\sigma)} > 1
}
The converse follows similarly, thus completing the proof.
\end{IEEEproof}
\begin{prop}[$\epsilon$-Synchronization]\label{propeps}
For a stationary ergodic  system $\pax \models \wax$, we have:
\cgather{
\forall \epsilon > 0, \exists x_\epsilon \in \Sigma^\star, \textrm{ s.t. } \exists q \in Q, Pr(q \vert x_\epsilon) \geqq 1 - \epsilon
}
\end{prop}
\begin{IEEEproof}
Assume, if possible that for some  $x' \in \Sigma^\star$, 
$\wp_{x'}= Pr(i \vert x')$, we have:
\cgather{\label{eqassume}
\left ( u \triangleq \sup_{j \in Q} \wp_{x'} \big \rvert_j  < 1 \right ) \bigwedge \left ( \forall x \in \Sigma^\star, \sup_{j \in Q} \wp_{x'x}  \big \rvert_j \leqq u \right )
}
First, we claim:
\cgather{\label{supclaim}\exists j \in Q, \textrm{ s.t. } \wp_{x'} \vert_j = u
}
$i.e.$, the supremum is achieved
by some state. This is trivially true if $\abs{Q}< \infty$. We claim, it is also true in the general countable case. To see this, note that if for some $i''$, we have:
\cgather{
\exists \ell_1,\cdots, \ell_r, \cdots \textrm{ s.t. } \wp_{i''} \geqq \cdots \geqq \wp_{\ell-1} \geqq \wp_{\ell} \geqq \cdots
\\\Rightarrow \forall N \in \mathbb{N}, \sum_{r=1}^N\wp_{\ell_r} \geqq \wp_{i''} N 
}
implying that for a countably infinite state space, where the supremum is never achieved, we must necessarily have $\wp_{i''}=0$,  resulting in contradiction, thus establishing Eq.~\eqref{supclaim}.

Now,  if $\exists \sigma \in \Sigma$ such that $\sup_{j \in Q} \wp_{x'\sigma}\vert_j$ is reduced below $u$, then there exists 
a symbol that increases it as well (Lemma~\ref{lembalance}). Hence, it follows that we must have:
\cgather{
\wp_{x'} \Pitilde = \phi_{x'} = \pitilde(j,\cdot)
}
and since the same argument applies for any extension of $x'$:
\cgather{
\forall x \in \Sigma^\star, \exists ! i \in Q, \textrm{ s.t. } \forall \sigma \in \Sigma, \phi_{x'x}\rvert_\sigma = \pitilde(i,\sigma)
}
Let us define:
\cgather{
L_i \triangleq \{ x \in \Sigma^\star:  \argsup_{j \in Q} \wp_{x'x}\big \rvert_j =i \}
}
It follows immediately:
\cgather{
\bigcup_{i\in Q} L_i = \Sigma^\star \textrm{ and } 
\forall i,j \in Q, L_i \bigcap_{i\neq j} L_j = \varnothing 
}
Clearly, we have the following bijections:
\cgather{
\zeta: \{ L_i\} \rightarrow Q, \\
\xi:\{L_i\} \rightarrow \mathbb{N}, \textrm{ s.t. } \xi(L_i) = i
}
We define a model $(\Sigma,Q^\Delta, \delta^\Delta, \pitilde^\Delta)$, such that:
\begin{subequations}
\cgather{
Q^\Delta = \{\xi(L_i)\}\\
\forall i \in Q^\Delta, \forall \sigma \in \Sigma, \delta^\Delta(i,\sigma) = \delta(\zeta\circ\xi^{-1}(i),\sigma) \\
\forall i \in Q^\Delta, \forall \sigma \in \Sigma, \pitilde^\Delta(i,\sigma) = \pitilde(\zeta\circ\xi^{-1}(i),\sigma) 
} 
\end{subequations}
Interpreting $Q^\Delta$ as a simple renaming of $Q$, we note that $\Sigma,Q^\Delta, \delta^\Delta, \pitilde^\Delta)$ is indistinguishable from $\pax$. Hence, comparing the equivalence class of $x'$ in the identical models:
\cgather{
\eqcl[x']^{\Delta} = \zeta(L_i) \Rightarrow \eqcl[x']=\xi\circ\zeta^{-1}\circ\zeta(L_i) = i \Rightarrow u=1
}
which contradicts Eq.~\eqref{eqassume}. Hence,  we have either $u=1$, or 
\cgather{
\exists  x \in \Sigma^\star, \sup_{j \in Q} \wp_{x'x}  \big \rvert_j > u
}
In either case, we have the desired result.
\end{IEEEproof}
\begin{cor}[To Proposition~\ref{propeps}: Joint $\epsilon$-synchronization]\label{corepsjoint}
Given two ergodic stationary systems, $G = \pax \models \wax$, and $G' = (\Sigma,Q',\delta',\pitilde') \models (\Sigma^\omega,\F,\mu')$,  we have:
\calign{
\exists x_\epsilon \in \Sigma^\star, & \textrm{ such that } 
 \left ( \exists q \in Q,   Pr(q \vert x_\epsilon) \geqq 1 - \epsilon \right )  \notag \\ & \mspace{50mu} \bigwedge
 \left ( \exists q' \in Q', Pr(q' \vert x_\epsilon) \geqq 1 - \epsilon\right )  
}
\end{cor}
\begin{IEEEproof}
We define  $G''=(\Sigma\times \Sigma,Q'',\delta'',\pitilde'')$:
\calign{
&Q'' \triangleq Q \times Q'\\
&\forall i \in Q,  j \in Q', \forall \sigma,\sigma' \in \Sigma, \notag \\  & \mspace{50mu} \left \{  \begin{array}{l}\delta''((i,j),(\sigma,\sigma')) \triangleq ( \delta(i,\sigma),\delta'(j,\sigma'))\\
\pitilde''((i,j),(\sigma,\sigma')) \triangleq \pitilde(i,\sigma) \pitilde'(j,\sigma') \end{array}\right.
}
It is easy to verify that:
\cgather{
\sum_{\mathclap{(\sigma,\sigma')\in\Sigma\times \Sigma}}  \pitilde''((i,j),(\sigma,\sigma')) = \sum_\sigma  \pitilde(i,\sigma) \sum_{\sigma'}\pitilde'(j,\sigma') = 1
}
implying that $G''$ is a valid model. Now applying Proposition~\ref{propeps}, we conclude that:
\cgathers{
\forall \epsilon > 0, \exists x_\epsilon \in \Sigma^\star, \textrm{ s.t. } \exists q \in Q, q'\in Q', Pr((q,q') \vert x_\epsilon) \geqq 1 - \epsilon
}
The absence of any interaction in the dynamics of $G,G'$ in the construction of $G''$, then implies that $x_\epsilon$ jointly $\epsilon$-synchronizes both $G,G'$. This completes the proof.
\end{IEEEproof}
%
\subsection[Vector Space of Ergodic Stationary Processes]{Vector Space of Ergodic Stationary  Processes}

\begin{defn}[Strictly Positive Ergodic Stationary Processes]\label{defposproc} A strictly positive process over a finite alphabet $\Sigma$ is a finite-valued  stationary ergodic process  such that:
\cgather{\label{eqproppos0}
\forall x \in \Sigma^\star, \forall \sigma \in \Sigma, \phi_x \big \rvert_\sigma > 0
}
$\Sp$ denotes the  space of positive processes over  $\Sigma$.
\end{defn}
Note that a finite valued stationary ergodic process is a positive process if and only if every symbolic derivative a strictly positive probability vector on $\Sigma$.

\begin{defn}[Scalar Product]\label{defscalarprod}  
For an ergodic stationary process $\pax \models \wax$, we can construct the scalar product $\alpha \odot \pax$ as follows: 

For a $\epsilon$-synchronizing string $x_\epsilon$, 
\cgather{
\forall x \in \Sigma^\star, \phi'_{x} \triangleq \lim_{\epsilon \rightarrow 0^+}\alpha \odot \phi_{x_\epsilon x} 
}
We note that:
\cgather{
\exists x_0 \in \Sigma^\star, \phi'_\lambda \longrightarrow \phi_{x_0} \\
\textrm{ where if } \eqcl[x_0]=i_0 \in Q, \textrm{ then } Pr(i_0 \vert x_\epsilon) \longrightarrow 1
}
We define a map $\zeta: \Sigma^\star \rightarrow Q$ as:
\cgather{
\zeta(\lambda) = i_0\\
\forall x \in \Sigma^\star, \sigma \in \Sigma, \zeta(x\sigma) = \delta(\zeta(x),\sigma)
}
Then, we  construct a model $G'=(\Sigma,Q',\delta',\pitilde')$ as:
\cgather{
Q'=Q\label{eqz1}\\
\forall x \in \Sigma^\star, \delta'(\zeta(x),\sigma) = \zeta(x \sigma) \label{eqz2}\\
\forall x \in \Sigma^\star, \pitilde'(\zeta(x), \cdot) = \phi'_x \label{eqz3}
}
Finally, we define:
\cgather{
\alpha \odot \pax = \CLX{G'}
}
\end{defn}
\begin{lem}[Scalar Product]\label{lemstochscalar}
The construction of $G'=(\Sigma,Q',\delta',\pitilde')$  in Definition~\ref{defscalarprod} is consistent.
\end{lem}
\begin{IEEEproof}
We only need to establish that $Q'=Q$ in Eq.~\eqref{eqz1} is consistent with the definition of 
$\delta', \pitilde'$ in Eqns.~\eqref{eqz2} and \eqref{eqz3}, which follows  from noting that $i_0 \in Q'$ since there exists some sequence $x\sigma$ such that $\zeta(x\sigma) = q_0$ since $\pax$ is a closed restriction. For the same reason, there exists  sequences beginning from $q_0$ visiting every state in $Q$, implying that if we construct $Q'$ using Eq.~\eqref{eqz2}, then we end up with $Q'=Q$. This completes the proof.
\end{IEEEproof}
\begin{defn}[Sum]\label{defstochsum}
For ergodic stationary processes $\pax \models \wax$, $\pax['] \models (\Sigma^\omega,\F,\mu')$, a  closed commutative binary operation $(\Sigma,Q'',\delta'',\pitilde'') = \pax \oplus \pax[']$ may be  constructed as follows: 

For a jointly $\epsilon$-synchronizing string $x_\epsilon$, 
\cgather{
\forall x \in \Sigma^\star, \phi''_{x} \triangleq \lim_{\epsilon \rightarrow 0^+}  \phi_{x_\epsilon x} \oplus \phi'_{x_\epsilon x} 
}
Denoting  state probabilities in $\pax[']$ as $Pr'(\cdot)$, we note:
\cgather{
\exists x_0 \in \Sigma^\star, \phi''_\lambda \longrightarrow \phi_{x_0}\oplus \phi'_{x_0} \\
\textrm{ where if } \eqcl[x_0]=i_0 \in Q, \eqcl[x_0]=j_0 \in Q' \notag \\ \textrm{ then } Pr(i_0 \vert x_\epsilon) \longrightarrow 1 \bigwedge  Pr'(j_0 \vert x_\epsilon) \longrightarrow 1
}
We define a map $\zeta: \Sigma^\star \rightarrow Q\times Q'$ as:
\cgather{
\zeta(\lambda) = (i_0,j_0)\\
\forall x \in \Sigma^\star, \sigma \in \Sigma, \zeta(x\sigma) = \delta(\zeta(x),\sigma)
}
Then, we  construct a model $G''=(\Sigma,Q'',\delta'',\pitilde'')$ as:
\cgather{
Q''=Q\times Q'\label{eqz1}\\
\forall x \in \Sigma^\star, \delta''(\zeta(x),\sigma) = \zeta(x \sigma) \label{eqz2}\\
\forall x \in \Sigma^\star, \pitilde''(\zeta(x), \cdot) = \phi''_x \label{eqz3}
}
Finally, we define:
\cgather{
G \oplus G' = \CLX{G''}
}
\end{defn}
\begin{lem}[Sum]\label{lemstochsum}
The construction of $G''=(\Sigma,Q'',\delta'',\pitilde'')$  in Definition~\ref{defscalarprod} is consistent.
\end{lem}
\begin{IEEEproof}
As in Lemma~\ref{lemstochscalar}, we only need to establish that $Q''=Q\times Q'$  is consistent with the definitions of 
$\delta'', \pitilde''$, which follows by beginning with $(i_0,j_0) \in Q''$, and recalling that both $\pax,\pax[']$ are closed restrictions.
\end{IEEEproof}

\begin{notn}As in the case of probability vectors,  we denote $\oplus,\odot$ in the context of processes as simply $+$ and concatenation, if no confusion arises.
\end{notn}
The commutative sum of stochastic processes established above induces an Abelian group on $\Sp$. We note that process equivalence (and uniqueness) is upto equality of finite dimensional distributions (FDD equivalence).
\begin{lem}[Abelian Group on Stochastic Processes]\label{lemabelstoch} 
\begin{subequations}\cgather{
\forall G,G' \in \Sp, G+G' \in \Sp \label{eq123a}\\
G + G' = G'+G\label{eq123b} \\
\exists ! \Zr \in \Sp, \textrm{ such that }\forall G \in \Sp,  G+\Zr=G\label{eq123c}\\
\forall G \in \Sp,\exists ! G' \in \Sp,  \textrm{ such that }  G+G'=\Zr\label{eq123d}
}
\end{subequations}
where uniqueness  is assumed upto FDD equivalence.
\end{lem}
\begin{IEEEproof}
Eqns.~\eqref{eq123a} and \eqref{eq123b} are immediate from Definition~\ref{defstochsum}.
Now, using the fact that a complete set of symbolic derivatives uniquely specifies a process upto FDD equivalence (Lemma~\ref{lemsuff}), we define a stationary ergodic process  $W$ as:
\cgather{
\forall x \in \Sigma^\star, \phi_x^{W} = \zr[\abs{\Sigma}]\label{eqW}
}
where $\zr[\abs{\Sigma}]$ is the uniform probability vector over $\Sigma$.
We claim:
\cgather
{
\forall G \in \Sp,  G+W=G \tag{Claim A}\\
\forall G \in \Sp,  G+H=G \Rightarrow H =W \tag{Claim B}
}
The first claim follows from noting that  for any $\epsilon$-synchronizing sequence $x_\epsilon$ for $G$,
 (using $\phi^G_x$ to denote the symbolic derivative for $G$ at $x$) we have:
\cgather{
\forall x \in \Sigma^\star, 
\phi^G_{x_\epsilon x} \oplus \zr[\abs{\Sigma}] = \phi^G_{x_\epsilon x} 
} 
For the second claim we begin by noting that if $G+H = G$ for all $G \in \Sp$, then, for any fixed  $G$,  we must have all the
finite dimensional distributions for $G+H$ and $G$ coincide, $i.e.$:
\cgather{
\forall x \in \Sigma^\star, \textrm{ s.t. }  \phi^{G+H}_x =  \phi^G_x \label{eq138}
}
Now, using the notation used in the construction of the sum $G+H$ in Definition~\ref{defstochsum}, we have:
\cgather{
 \forall x \in \Sigma^\star, \phi''_{x} = \lim_{\epsilon \rightarrow 0^+}  \phi^G_{x_\epsilon x} \oplus \phi^H_{x_\epsilon x} 
}
where $\forall \epsilon > 0, x_\epsilon$ is a jointly $\epsilon$-synchronizing string. If $\pitilde^G, \pitilde^H, \pitilde^{G+H}$ are the morph matrices, and $Q^G, Q^H, Q^{G+h}$ are the state sets for $G,H,G+H$ respectively, it follows that:
\mltlne{
\forall q \in Q^{G+H}, \exists q_G \in Q^G, q_H \in Q^H, \\
\pitilde^{G+H}(q, \cdot) = \pitilde^G(q_G, \cdot) +\pitilde^H(q_H, \cdot)
}
Since we necessarily have (Proposition~\ref{propeps}):
\cgather{
\forall q \in Q^{G+H}, \forall \epsilon > 0,  \exists x_\epsilon \in \Sigma^\star,  Pr^{G+H}(q \vert x_\epsilon) > 1 -\epsilon
}
it follows that:
\mltlne{
\forall q \in Q^{G+H},  \forall \epsilon > 0, \exists x_\epsilon \in \Sigma^\star,\\ \norm{\phi^{G+H}_{x_\epsilon} - \pitilde^{G+H}(q, \cdot) } < \epsilon
}
which then implies from Eq.~\eqref{eq138}:
\cgather{
 \exists q_G \in Q^G, q_H \in Q^H,\pitilde^G(q_G, \cdot) +\pitilde^H(q_H, \cdot) = \pitilde^G(q_G, \cdot) \\
\Rightarrow \exists q_H \in Q^H, \pitilde^H(q_H, \cdot) = \zr[\abs{\Sigma}]
}
We recall that  the unique minimal closed restriction  operation in the last step
of the construction described in Definition~\ref{defstochsum} implies $Q^{G+H} \subseteqq Q^G \times Q^H$. However,
we cannot eliminate any $q_H \in Q^H$ completely from  the Cartesian product, $i.e.$:
\cgather{
Q^{G+H} \nsubseteqq Q^G \times \left ( Q^H\setminus {q_H} \right ) 
}
which follows from the fact that we assume all models to be minimal closed restrictions.
Hence, it follows that:
\cgather{
\forall  q_H \in  Q^H, \pitilde^H(q_H, \cdot) = \zr[\abs{\Sigma}]
}
implying that in the process modeled by $H$, all 
sequences are equivalent, with the symbolic derivatives as given in Eq.~\eqref{eqW}.
This establishes Claim B, and establishes Eq.~\eqref{eq123c}, where the required $\Zr$ is given by $W$.

To establish Eq.~\eqref{eq123d}, given $G=\pax$, we construct $G'=(\Sigma,Q,\delta,\pitilde)$ as:
\cgather{
\forall q \in Q, \pitilde'(q,\cdot) = -1 \times \pitilde'(q,\cdot)
\intertext{
and claim that:}
G + G' = \Zr
}
which follows from noting that (using the notation of  Definition~\ref{defstochsum}), we have:
\cgather{
 \forall x \in \Sigma^\star, \phi''_{x} = \lim_{\epsilon \rightarrow 0^+}  \phi^G_{x_\epsilon x} \oplus \phi^{G'}_{x_\epsilon x} = \zr[\abs{\Sigma}]
}
Uniqueness of $G'$ follows from:
\cgather{
G + G' = G+ G'' = \Zr \Rightarrow G' = G''
}
This completes the proof.
\end{IEEEproof}

\begin{lem}[Vector Space]\label{lemvecsp}
$\Sp$ satisfies the following:
\begin{subequations}\cgather{
\forall \alpha \in \mathbb{R}, \forall G \in \Sp, \alpha G \in \Sp \\
\forall G \in \Sp, 0G = \Zr \\
\forall \alpha \in \mathbb{R}, \forall G,G' \in \Sp, \alpha (G+G') = \alpha G + \alpha G'\\
\forall \alpha \in \mathbb{R}, \forall G \in \Sp, \alpha G + (-\alpha) G = \Zr\\
\forall \alpha,\alpha' \in \mathbb{R}, \forall G \in \Sp, (\alpha\alpha')G = \alpha (\alpha' G) = \alpha' (\alpha G)
}
\end{subequations}
\end{lem}
\begin{IEEEproof}
Immediate from Definition~\ref{defscalarprod}, and corresponding definitions for probability vectors.
\end{IEEEproof}

\section{Inner Product of Ergodic Stationary   Processes}

\begin{defn}[Inner Product of Stochastic Processes]\label{defipstoch}
For a strictly positive ergodic stationary processes $\pax \models \wax$, $\pax['] \models (\Sigma^\omega,\F,\mu')$,
\cgather{
\langle G, G'\rangle \triangleq \lim_{\epsilon \rightarrow 0^+} \lim_{N\rightarrow \infty} \frac{1}{N} \sum_{i=0}^{N} \langle \phi^G_{x_\epsilon x_i},\phi^{G'}_{x_\epsilon x_i} \rangle
}
where $\forall \epsilon > 0, x_\epsilon$ is a  jointly $\epsilon$-synchronizing sequence, and $\forall i \in \mathbb{N}, x_0 = \lambda, x_{i} = x_{i-1}\sigma$ with $\sigma$ drawn uniformly from $\Sigma$.
\end{defn}
Note that if $\abs{Q}=1,\abs{Q'}=1$, then $\langle G, G'\rangle$ is indeed a valid inner product, based on the formulation of inner products 
on finite dimensional probability vectors in Section~\ref{secippropbvec}. Thus, for strictly positive i.i.d. processes taking values over a finite alphabet, 
we have a valid inner product. In general, we have:
\begin{lem}[Complete Inner Product]\label{lemconsistency}
Definition~\ref{defipstoch} defines a complete inner product on the space of strictly positive stationary ergodic finite-valued processes.
\end{lem}
\begin{IEEEproof} (Sketch, details omitted.)
Since $\langle \phi_{x_\epsilon x_i},\phi'_{x_\epsilon x_i} \rangle$ is a valid inner product on $\Pp[\abs{\Sigma}]$, and noting that joint synchronization extends to a finite number of sequences (and hence we can find a jointly $\epsilon$-synchronizing sequence for any triplet of ergodic stationary processes $G,G',G''$), we conclude that:
\mltlne{
\forall \alpha \in \mathbb{R}, \langle G, \alpha (G'+G'') \rangle = \alpha \langle G, G' +G''\rangle   \\= \alpha \left ( \langle  G, G' \rangle  +\langle G, G'' \rangle \right )
}
Symmetry and non-negativity is also immediate. 
%
%
To prove completeness, we need to show that any Cauchy sequence in the space of
our class of stochastic processes converges within our class. This is immediate since, if any sequence of processes converges 
outside our class,  then the norm of the limiting process increases without bound,
implying the sequence is not Cauchy (by the same argument used in Lemma~\ref{lemcauchy}).
\end{IEEEproof}

\section{Example}
We consider a simple example of the resilience of the inner product to noise corruption. We consider two processes generated by two state PFSAs, and hence are infact ergodic and stationary (See Fig.~\ref{figex1}). The noise corrupted versions are shown as well. The uncorrupted processes are easy to distinguish, while post-corruption it becomes a difficult problem to discriminate them from each other, as well as from the average iid approximation. A simple calculation shows that the relative angles remain mostly unchanged, which suggests a new approach to process classification/discrimination in high noise scenarios. Plate F in Fig.~\ref{figex1} shows the separation achieved using computation of relative angles from corrupted data-streams (the generated binary data streams have means $0.499$ and $0.5004$, and standard deviations of $0.5$, suggesting that they are indeed very close to flat white noise. Detailed comparison with standard techniques is being carried out at present. 

\section{Summary, Conclusion \& Future Work}
We developed a Hilbert space for ergodic stationary processes, which would potentially allow us to investigate intrinsic structure of data in high noise scenarios. Future work will pursue detailed comparison with state of the art, and explore classification and clustering strategies based on the theoretical foundation developed here.

\begin{figure}[t]
\centering

\includegraphics[width=3.5in]{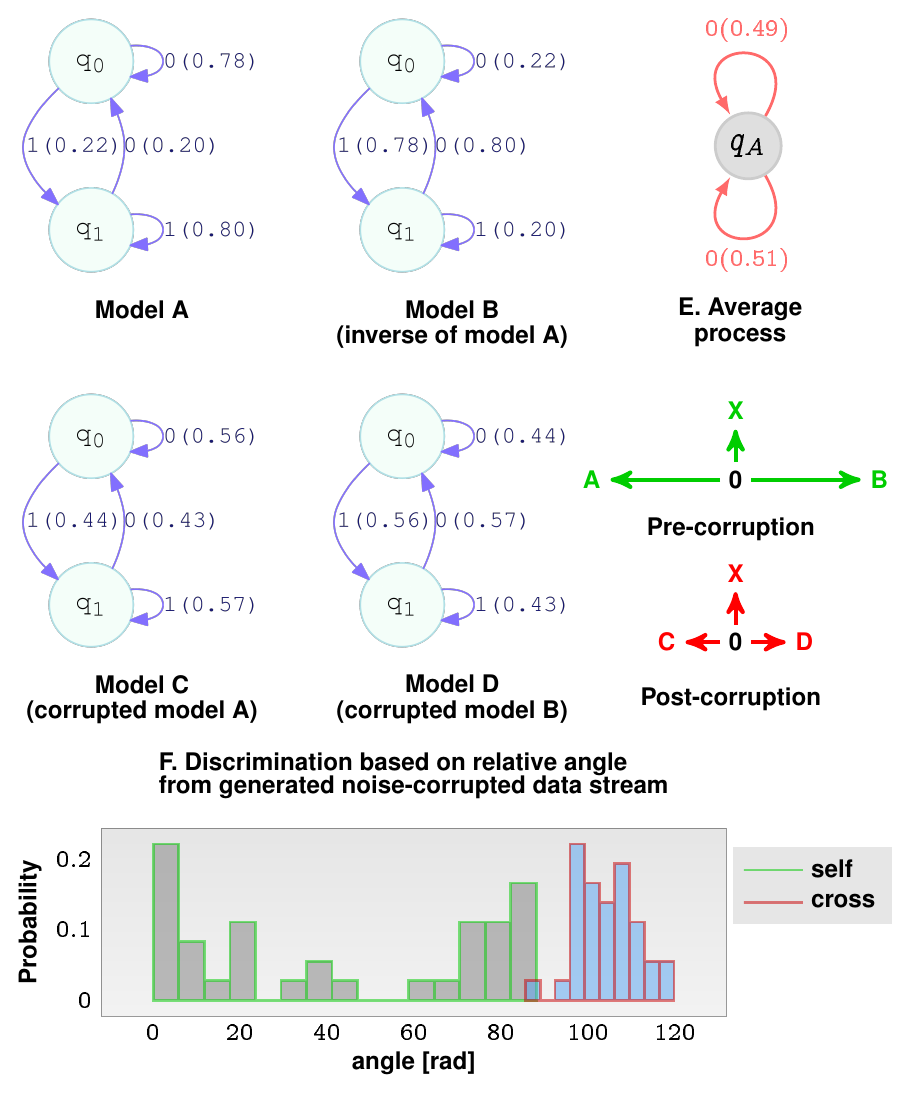}

\captionN{Typical example: We consider a process generated by a 2 state PFSA $G$ (model A). Model B is a realization of $-G$, and hence the angle between A and B is $\pi$ radians. Model C is a severely corrupted version of $G$, obtained as $0.1G$ (Note that $0\times G$ is flat white noise). Model D is $-0.1G$, or an equally corrupted version of model B. Importantly, it is easy to discriminate models A and B, particularly from the average or the iid approximation shown as model X. Noise corruption however makes it very difficult to disambiguate the processes generated by models C and D from each other, and particularly from that generated by model X. The angles between the processes are shown in middle right, where it is shown that for unstructured noise corruption, the relative angles are almost invariant. The key point here is that noise reduces the distance between the processes, but, under weak assumptions, leaves the intrinsic geometry relatively unchanged. Plate F shows the discrimination achieved with generated data. Abscissa shows the angle ``from self'' (ideally zero), and that between the processes (ideally $\pi$)  }\label{figex1}

\end{figure}
}

\bibliographystyle{siam}
\bibliography{BibLib1}


\end{document}